# Are Current Continual Learning Methods Truly Agnostic? Introducing OPRE, a Step Toward Agnostic Continual Learning

**Raphaël Bayle[1] (Corresponding author), Martial Mermillod[1], and Robert M. French[2]**

[1] Univ. Grenoble Alpes, Univ. Savoie Mont Blanc, CNRS, LPNC, Grenoble, France

[2] LEAD-CNRS UMR 5022, U. de Bourgogne Europe, Dijon, France

**Abstract**

In order to achieve Continual Learning (CL), the problem of catastrophic forgetting, one that has plagued neural networks since their inception, must be overcome. The evaluation of continual learning methods relies on splitting a known homogeneous dataset and learning the associated tasks one after the other. We argue that most CL methods introduce a priori information about the data to come and cannot be considered agnostic. We exemplify this point with the case of methods relying on pretrained feature extractors, which are still used in CL. After showing that pretrained feature extractors imply a loss of generality with respect to the data that can be learned by the model, we then discuss other kinds of a priori information introduced in other CL methods. We then present the Online Patch Redundancy Eliminator (OPRE), an online dataset-compression algorithm that discards information through two explicit, input-space criteria. With a classifier that was randomly initialized at test time, OPRE's performance matches reported state-of-the-art online continual-learning methods on CIFAR-10 and CIFAR-100 without any pretrained feature extractor, and outperforms GDumb at an identical memory budget—while making only minimal and interpretable assumptions about the data to come. We frame these results as an empirical, information-theoretic perspective on continual learning.

## I. Introduction

A key feature of intelligence is adaption to change. Nevertheless, after their initial training, artificial neural networks fail to learn new data without forgetting what they previously learned: this problem has been described in the late eighties and named *catastrophic forgetting* (CF) (French, 1999). This issue is addressed by the field of continual learning, also named incremental learning or lifelong learning, in which methods enabling artificial neural networks to avoid forgetting old tasks when learning new tasks have been developed. This field is grouping different kinds of methods, often split into three categories: regularization-based methods, architecture-based methods and replay methods. Regularization methods aim at reducing changes in the neural-network weights which are the most useful with respect to previously learned tasks (Aljundi et al., 2018; Kirkpatrick et al., 2017; Zenke et al., 2017). Architecture-based methods increase the size of the neural network to learn new tasks without affecting the part of the network dedicated to formerly learned tasks (Yan et al., 2021; Yoon et al., 2017). Replay and pseudo-replay based methods rely on the storage or generation of examples seen during past tasks (or pseudo-examples based on prior learning), to be inserted with the new data to be learned (Buzzega et al., 2020; Robins, 1995).

Various scenarios are studied in continual learning. Their difficulty varies, in particular, depending on whether task labels are available or not at test time. In this work, the focus is on class-incremental learning, where new tasks contain data related to new classes and the identity of the task to which a sample belongs is not provided at test time, according to the definition of van de Ven et al (Van De Ven et al., 2022). Typically, in this scenario, a dataset is split into tasks that contain every item of a selected subgroup of N classes, and, at test time, only samples without further information (such as the task to which the sample belongs) are provided. Online continual learning introduces an additional constraint by assuming that training exemplars are provided one at a time (Aljundi, 2019).

Despite many methods addressing the challenge of task agnostic class-incremental continual learning, due to the relative homogeneity of all the data belonging to a given dataset, before it is split, some of those methods are not totally agnostic, in the sense that they contain an implicit hypothesis about data to come. This is especially the case for models using a pretrained feature extractor: such a feature extractor will render salient a limited number of features of the input and ignore others, based on the task and data used for pretraining. Pretrained feature extractors are still used in recent methods (Hayes & Kanan, 2020; Pourcel et al., 2022; Shi et al., 2025). We think it is worth discussing this point in detail, as it may guide the development of future methods.

The properties that we think are desirable for continual learning methods for a task of classification are as follows:

1. The method should be able to retain information related to previously seen tasks that are no longer available,
2. The method should be able to adapt to new tasks,
3. The method should use as few hyperparameters as possible, and those hyperparameters must be interpretable, i.e. having a “physical” meaning in relation to the input space,
4. The method should adapt its memory requirements to the complexity of the data.

It is to be noted that this fourth point is debated in the literature. For instance, Peng & Risteski (2022) considers that continual learning (unlike lifelong learning) is about updating a fixed-size model, whereas Yan et al. (2021) considers that a continual-learning algorithm can involve increasing the size of a network as new tasks are encountered. We argue that, if no information is known about the data to come, memory consumption must be able to adapt dynamically to the data: if the size of a particular model is too small, then it will be unable to learn large data sets (assuming the data are complex enough). If the size of the model is too large, a significant proportion of its memory will be unnecessary.

In this work, we refer to *agnostic continual learning* as continual learning methods that rely on no prior assumptions about the structure of future data. Strictly speaking, any choice of hyperparameters, pretrained parameters, or model architecture can be interpreted as prior information, since these choices are typically optimized to perform well on an entire sequence of tasks—even when the data from these tasks are not available simultaneously. Agnostic continual learning should be evaluated on classification problems generated from arbitrary class definitions over the input space. Such an evaluation is, however, infeasible in high-dimensional settings, and standard CL benchmarks—built from split homogeneous datasets—cannot meaningfully test such a property.

Because ideal agnostic continual learning cannot be implemented or evaluated in practice, we instead study what we consider a meaningful step toward it: a continual learning method whose only bias arises from explicit and interpretable data simplification, i.e., a controlled and transparent loss of information. This perspective is particularly relevant in settings where future data may come from distributions that differ radically from those seen so far. Under this principle, if the discriminative structure of a task survives the simplification, learning remains possible. In this direction, we propose OPRE, which uses only simple hyperparameters defined directly in the input space—namely, a pixel-level discretization and a minimum Euclidean distance between patches. These assumptions make the behavior of the algorithm transparent and avoid introducing hidden priors about the structure of future tasks.

Agnostic continual learning, as defined above, should not be confused with task-agnostic continual learning (Zeno et al., 2018). Task-agnostic methods assume that task boundaries are unknown or not provided, but they still rely on the implicit structure of the underlying benchmark: the data are drawn from a fixed sequence of tasks, typically sharing a common modality, resolution, and semantic organization. In other words, task-agnostic CL removes access to task labels but preserves strong assumptions about how tasks are generated. In contrast, our notion of agnostic continual learning removes assumptions at a deeper level: not only are task boundaries unknown, but the very nature of future tasks is unconstrained. Future data may correspond to arbitrary class definitions over the input space, potentially exhibiting extreme distribution shifts. This distinction is crucial: task-agnostic CL addresses uncertainty about when tasks change, whereas agnostic CL addresses uncertainty about what future tasks could be.

Our work is divided in two parts. In the first part, we study the use of pretrained feature extractors. We propose two simple experiments. The first aims at evaluating the extent to which pretrained feature extractors can solve the problem of catastrophic forgetting on commonly used datasets with minimal additional algorithms. Specifically, we tested the prediction-performance of a pretrained feature extractor with a simple nearest-class mean classifier. We show that this very simple classification algorithm, which can be used as a continual-learning method, yields near the state-of-the-art performance, indicating that the representations obtained with the feature extractor are much easier to classify than their associated raw images. The second experiment aims at showing the cost of using a pretrained feature extractor. We demonstrate that extracted features of data that are far from those used for during pretraining result in the impossibility of classifying them.

In the second part of our work, we propose a simple alternative to feature extraction that solves the problem of implicit information deletion due to feature extraction. We propose an algorithm that we have called the Online Patch Redundancy Eliminator (OPRE), an online dataset compression algorithm designed to store the diversity of incoming data efficiently by identifying and removing redundancies according to an explicit criterion. We motivate the interest of continual learning through dataset compression by examining flaws in other methods. Most methods, even if they do not rely on a pretrained feature extractor, either i) rely on a fixed-size architecture which is calibrated to the size and complexity of the whole dataset from which the tasks are derived and cannot be considered agnostic as such or ii) rely on building new fixed-size blocks for each new task, regardless of the amount of novelty introduced by the new tasks (Yan et al., 2021; Zhang et al., 2023). For this latter category of methods, even though pruning strategies reduce the size of the model depending on the complexity of learnt data, they rely on additional hyperparameters, require additional computations and are sensitive to the size of tasks. They also require the storage of some raw data for each of the previously seen classes (Yan et al., 2021). Our algorithm is designed to dynamically identify redundancies at the level of small square “patches” of each image between new images and previously seen images, thereby reducing the memory requirements for the new images. The compression is lossy with a simple, explicit criterion – namely, a minimal Euclidian-distance threshold for each new patch with respect to previously stored patches -- to decide whether the new patch must be kept or discarded. Pixel values are also quantized: each new patch is discretized as soon as it arrives, before being compared with the previously stored (themselves discretized) patches. Classification is performed at test time by training a convolutional neural network on the stored compressed data, as in GDumb (Prabhu et al., 2020). We evaluate OPRE on CIFAR-10 and CIFAR-100, where it matches or surpasses reported state-of-the-art online continual-learning methods without any pretrained feature extractor and outperforms GDumb at an identical memory budget, while still being able to learn the synthetic data from the first part of this work.

# II. Related work

In this section, we categorize different methods of overcoming the problem of catastrophic forgetting with respect to how agnostic those methods are (i.e., how little or much they rely on prior information). In addition, we present and discuss existing methods linked to dataset compression and their potential link with continual learning.

## 1. Continual learning (CL)

***Pretrained feature extractor***

Several CL methods rely on a pretrained feature extractor with frozen weights. Deep Streaming Linear Discriminant Analysis (Hayes & Kanan, 2020), for instance, continuously trains an LDA classifier using the extracted feature vector from a large model pretrained on ImageNet. REMIND (Hayes et al., 2020) is an approach with a pretrained feature extractor that is trained on a fraction of the dataset and whose parameters are then frozen. The extracted low-dimensional features are then stored and replayed to avoid CF. This approach implicitly assumes that the relevant features of the first task are the same as those of the following tasks. Dynamic Sparse Distributed Memory (DSDM; Pourcel et al., 2022) is a dynamic variant of Sparse Distributed Memory (Kanerva, 1988; Flynn et al., 1987; Bricken et al., 2023), an associative memory that stores and retrieves high-dimensional vectors through a set of sparsely activated addresses—here created incrementally as new data arrive rather than fixed in advance, so that the memory grows with the incoming stream. In the work of Pourcel et al., DSDM operates not on raw images but on the features extracted by a pretrained feature extractor. While these methods require the feature extractor to be stored, the memory overhead of new data is very small, this being a significant advantage. Regardless of how agnostic they are, another advantage is that they perform in an online manner, meaning that samples can arrive one by one and that, at any time, the model can perform a classification. More recently, Shi et al. (2025) proposed a bioinspired approach to CF that relies on a feature extractor from a pretrained CLIP model.

***Continuously trained feature extractors: representation learning***

Other methods involve continuously training a feature extractor. ICaRL (Rebuffi et al., 2017) and Co2L (Cha et al., 2021) are two well-known examples of this kind of method. ICaRL relies on continuously training a feature extractor using a small buffer that stores exemplars from old tasks and then performing classification with a nearest-mean classifier. Co2L has a similar principle, but uses contrastive loss that enables learning richer representations. The main drawbacks of these methods are that the networks used are of a fixed size, meaning that they have a theoretically bounded learning capacity and that a buffer of raw exemplars is still required. Further, the size of the buffer is highly related to the model's performance, meaning that storing a diversity of raw examples is extremely important to a high-level of performance.

Other methods increase the size of their network to train the feature extractor, freezing older weights, and training on the new weights until the task is done. Their learning capacity is, thus, theoretically unlimited, but the number of parameters depend on the task structure and is independent of task difficulty, leading to a suboptimal use of parameters (Rusu et al., 2022). Progressive Neural Networks (Rusu et al., 2022) and the Adapter-based Continual Learning framework (Zhang et al., 2023) fall into this category of model. A drawback of these two methods is that they require the exemplars to be associated with their task identity at test time.

**Transformer-based methods**

Several attempts to adapt the Transformer architecture (Vaswani et al., 2017) to avoid catastrophic forgetting have been carried out. Among them, DyTox (Douillard et al., 2022) and Online-LoRA (Wei et al., 2024) have shown promising results. Online-LoRA continuously trains the model and adds new temporary parameters each time loss increases above a threshold; when new trainable parameters are added, they are merged with older frozen parameters and then frozen. This enables the network to continuously adapt to new data with a constant number of parameters. Initially the model is initialized with pretrained parameters on ImageNet-1K. In addition, Online-LoRA maintains a small buffer of raw training exemplars.

DyTox (Douillard et al., 2022) makes use of a continuously trained encoder along with a dynamic task decoder for a very low memory cost; the size of the encoder remains the same throughout training. A buffer containing samples from past tasks is also used. Both Online-LoRA and DyTox are sensitive to task size: for a fixed-size split dataset, the smaller the tasks, the lower the final classification accuracy.

Transformers can also be used for CL involving prompt learning (Z. Wang et al., 2022). In this method, prompts, used as input to a frozen pretrained transformer model, are learned dynamically. This method strongly relies on a pretrained network and is thus dependent on the knowledge learned by this pretrained model.

Most CL methods use a fixed quantity of memory, typically by using a fixed-size network which is updated during the continual learning process. We argue that this view of CL is of limited practical importance because it requires assumptions about the complexity of the data to come. The number of trainable parameters is directly linked to the complexity of the data a model can handle. In other words, either the complexity of the data on which the model is trained is low with respect to the number of the model’s parameters, or the complexity of the data is too high for the model size. In either case, the model will necessarily perform poorly. Model expansion, as proposed in DER (Yan, 2021), solves the issue of fixed size, but DER is not an online model and has performance strongly correlated with task size and relies on a buffer of raw exemplars. Online compression of all data on which the model is continuously trained could be a way to avoid being restricted by a fixed architecture or storing unnecessary data.

**Online continual compression.**

Most closely related to our approach is the work of Caccia et al. (2020), who introduced the problem of *online continual compression*: learning to compress and store a representative subset of a non-i.i.d. stream while observing each sample only once, under a fixed memory budget that must also accommodate the compressor itself. Their Adaptive Quantization Modules (AQM) build on the VQ-VAE, a learned discrete autoencoder whose codebook quantizes encoder outputs through a nearest-neighbor lookup, so that only the codebook indices need to be stored. AQM stacks several such modules with an adaptive multi-level storage scheme and an internal self-replay mechanism, together with dedicated machinery—codebook freezing, decoupled module training, and reservoir-style stream sampling—to counter the representation drift and forgetting that affect the compressor as it is trained online. Requiring no pretraining, AQM attains strong results on online continual-learning benchmarks and scales to higher-resolution images, LiDAR, and reinforcement-learning observations.

OPRE can be viewed as the non-parametric, training-free counterpart of this line of work. Both methods are, at heart, forms of online vector quantization; what differs is where the codebook comes from. AQM *learns* its codebook in a latent space optimized for reconstruction, whereas the "codebook" of OPRE is simply the set of discretized input-space patches it retains, greedily assembled into an ε-cover of the discretized patch space under the Euclidean metric directly in the input space. Consequently, OPRE introduces no learned compressor—and hence none of its failure modes: no representation drift, no compressor forgetting, and no autoencoder architecture to design—while its only information loss is governed by two interpretable hyperparameters defined directly on the input (the distance threshold ε and the discretization level), rather than by a reconstruction error measured in a learned, data-dependent representation. This is in line with our broader argument that an agnostic method should operate on the input data itself rather than on a transformed representation of it.

### 2. Dataset compression and continual learning

Deep Learning requires the use of huge amounts of data and, as a consequence, the field of dataset compression has emerged to try to reduce the size of training datasets (Comeau et al., 2025; Xiao et al., 2025). Some methods have, indeed, been successful (Zhou et al., 2023), but rely on intermediate models that are pre-trained on a fully known dataset, which is inappropriate for a continual-learning framework. Interestingly, (Xiao et al., 2025) show that methods relying on dataset distillation – i.e. keeping the soft labels provided by a pretrained model – perform less well than their approach based on maintaining diversity at the image level, without relying on soft labels.

Associating dataset compression with continual learning is not new: JPEG compression (Ahmed et al., 1974) has been used to reduce the memory cost of storing previous task exemplars (L. Wang et al., 2022), but setting the compression ratio is complicated. Additionally, JPEG-based dataset compression does not reduce the redundancy between the incoming data exemplars: two images that are almost identical will both require the same amount of memory.

## III. Feature extraction in continual learning

We now present two experiments designed to study the impact of using a pretrained feature extractor on data categorization. We will estimate the extent to which extracted features are simpler to categorize than raw exemplars. We will estimate the loss of generality caused by the use of a pretrained feature extractor by constructing a counterexample where feature extraction prevents correct categorization. Finally, we discuss the challenges associated with continuously training a feature extractor.

### 1. A cheap and efficient way to do continual learning using a pretrained feature extractor

In a multi-layered pretrained feature extractor, the features that are extracted are those produced by the layer just before a simple linear classifier. We can, therefore, expect that the extracted features to be well structured and easy to classify. To measure the extent of the simplification of the raw images done by the feature extractor, we tested a very simple classification method using extracted features – namely, classification based on the nearest class average. For every class, the feature vectors are accumulated and averaged, then the prediction for a test item is based on the class whose average feature vector has the smallest distance to the test vector. Note that because there is no interaction between class averages since each class has its own mean feature vector, the final accuracy in the class incremental paradigm is deterministic, independent of the order of presentation of the data and independent of how data are split into different tasks.

The performance of this nearest-class-mean classification for CIFAR-10, CIFAR-100 and CORe50 is presented in Table 1 and compared to DSDM (Pourcel et al., 2022), which relies on the same feature extractor, a ResNet-18 trained on ImageNet-1K. Despite its simplicity, the nearest-class-mean classifier using the extracted-feature vectors compares well with DSDM: it slightly outperforms DSDM on CIFAR-10, is outperformed by DSDM on CIFAR-100, but significantly outperforms DSDM on CoRE-50.

*Table 1. Comparison of simple nearest-mean classifier based on extracted features from a ResNet18 and DSDM*

| Dataset | Final Accuracy – Nearest-mean classifier (%) | Final Accuracy – DSDM, Best hyperparameters (%) |
|---|---|---|
| CIFAR-10 | 77.20 | 76.0 |
| CIFAR-100 | 53.52 | 63.3 |
| CoRE50 | 65.74 | 57.61 |

## 2. Limitations of the use of a pretrained feature extractor

We also want to show the limitations of using a pretrained feature extractor. Intuitively, a feature extractor is a function that reduces the dimensionality of inputs existing in a given space. Such a reduction is typically carried out by training a large neural network on a complex classification task and removing the final "classification" layer of the network. For example, a common feature extractor is ResNet-18, pretrained on ImageNet1k, with its final classification layer removed (Pourcel et al., 2022).

Obviously, the input data and the tasks used to train the feature extractor determine how the feature extraction is carried out. In the example of pretrained model obtained from a training on ImageNet1K, it seems reasonable that the extracted features would be relevant to categorize pictures from the real world, such as the ones in CoRE50, CIFAR-10, CIFAR-100 datasets. Nevertheless, using features derived from the original images implies a loss of generality and a model relying on a feature extractor should not be able to classify data belonging to other distributions, for example, images that are not "real-world images".

To prove this point, we carried out a simple experiment. We attempted to train a network to categorize based on the features extracted from two sets of artificial images separated by an arbitrary hyperplane. A fixed random vector $v_{hyperplane}$ is generated that is orthogonal to the separating hyperplane. Images with normally distributed pixels are generated and labeled 0 or 1 depending on the sign of the scalar product of their flatten representation with $v_{hyperplane}$. We show that, while a simple CNN or logistic regressor can correctly perform the classification on the raw images, it is not possible to perform the classification using the extracted features obtained from ResNet-18 pretrained on ImageNet1K, whether we use a one- or two-layer fully connected classifier. The details of this experiment are given in Appendix B.

## 3. Obstacles to continuously train a feature extractor

Another issue linked with extracting features, and that may arise even when trying to continuously update a feature extractor based on a classification task, is that the features learned from a given task (i.e. from a given set of exemplars available at the same time) may well differentiate the items in the current task, but are not necessarily representative of the classes themselves. For example, if a task is composed of images of white horses and white unicorns, the only differentiating feature is the presence of a horn; everything else is irrelevant. Thus, of all of the representational information used by the classifier at this time (legs, mane, size, color, presence of a tail, ears, etc.), only the presence or absence of horn will determine whether a given item is a unicorn or a horse.

Conversely, features can be extracted that are, in fact, irrelevant to the categorization but are used by the network to categorize. For example, a classifier might correctly classify birds and cars by relying on features such as the blue background (sky) for birds, and the ground beneath the car, neither of which are relevant features of birds or cars. A classic example of this can be found in Landecker et al. (2013).

Thus, extracted features depend not only on the data related to a given task, but also to other data in the task in which this class is learned. We suggest that this is why, to be able to extract relevant features, one must keep items from previous tasks. In almost all state-of-the-art training methods a buffer with raw exemplars is maintained (Cha et al., 2021; Douillard et al., 2022; Rusu et al., 2022; Wei et al., 2024; Zhang et al., 2023). Performance is generally a function of the buffer size (Cha et al., 2021; Rebuffi et al., 2017). This is still an issue when using a contrastive loss function, as in Co2L (Cha et al., 2021), which allows extracting richer representations.

To visualize how other data in the task influence the learning a particular class, we conducted the following experiment with the CIFAR-10 dataset. We compared the gradients for two cases when the input was airplanes. In one case, airplanes were learned along with birds and this was compared to a second case where airplanes were learned along with frogs. The gradients are displayed in Figure 1. It can be observed that the steepest gradient areas differ in the two conditions, indicating that classification is based on different parts of the images, thus on different features, depending on whether the network has been trained along with birds or with frogs. The network used in this experiment is described in Appendix A.

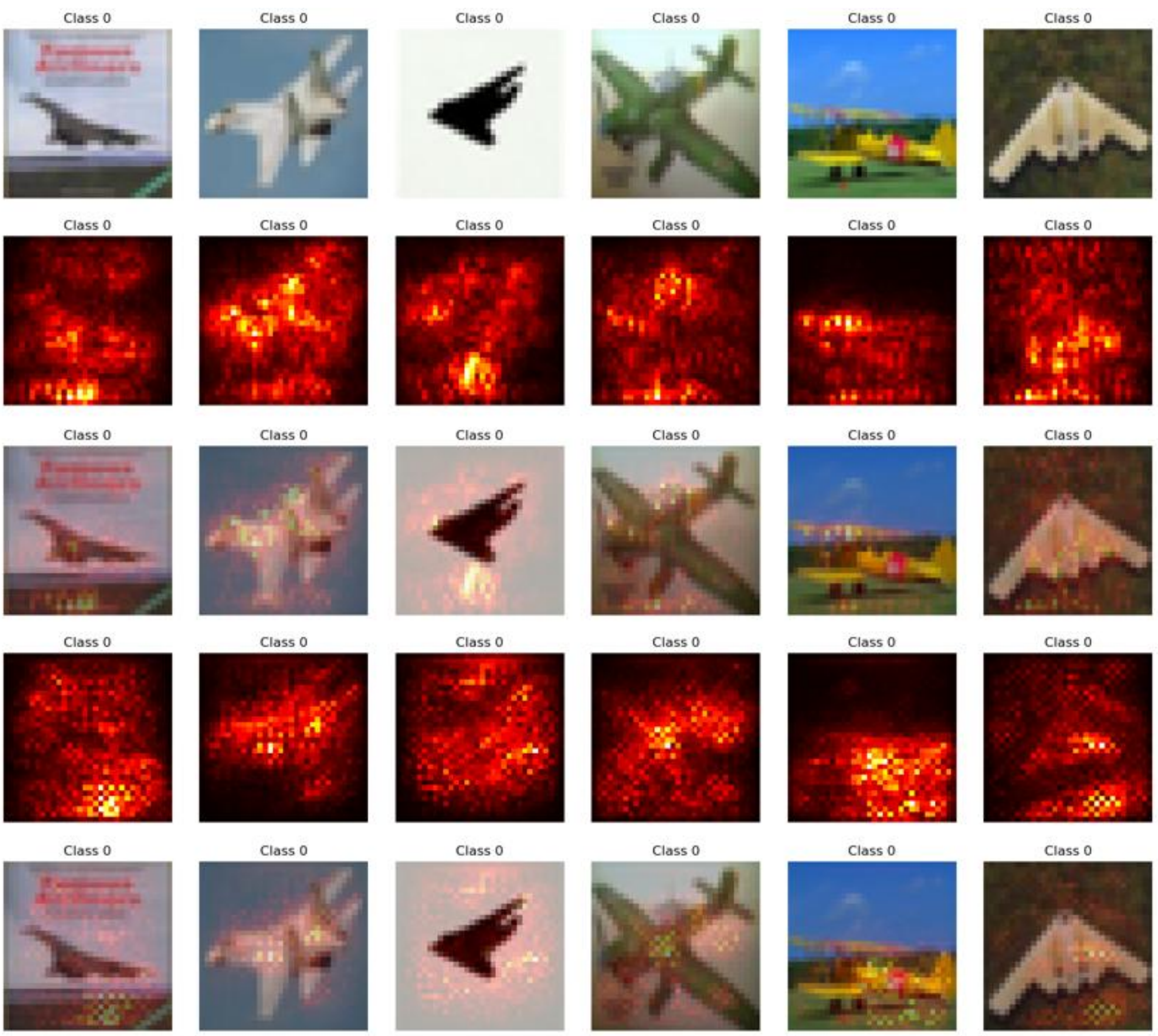

*Figure 1: First row: original picture; second and third rows: gradient and gradient superimposed on original picture after learning training set of birds and airplanes; fourth and fifth rows: gradient and gradient superimposed on original picture after*

*learning frogs and airplanes. Gradients are averaged over 5 runs, 10 epochs per run. Classification accuracy: birds+planes: 0.9202; frog+planes: 0.943*. Classification in the two cases is based on different extracted features.

### 4. Discussion

As previously shown, features extracted from commonly used datasets that are used to study continual learning, such as CIFAR-10, CIFAR-100 and CoRE50, can be surprisingly well classified with a very simple algorithm that selects the class whose mean feature vector is the closest to the one of a test image. This suggests that using a frozen pretrained feature extractor can be used to get a representation that captures the difference between the classes and that classification performance can be attributed to a significant degree to the feature extractor, independent of the details of the chosen continual-learning algorithm. We have also shown that using such a pretrained extractor with frozen weights comes at a cost – namely, that the features extracted from randomly generated, linearly separated classes cannot be classified using a Multilayer Perceptron (MLP), suggesting that a significant part of classification information is lost during feature extraction.

This also suggests that continual-learning algorithms based on the use of feature extractors implicitly make assumptions about the distribution of their input data. In practice, real world images have certain regularities that are captured by ResNet18 trained on ImageNet1K. The regularities of real-world images are well known. In particular, they have 1/f frequency distributions (Field, 1987). Our experiments suggest that using a feature extractor from a pretrained ResNet on ImageNet1K would be an extremely effective mean of classifying data that shares properties with the images in imageNet1K but would not be able to learn other data that do not share these properties. This suggests that CL algorithms that rely on a pretrained feature extractor are not agnostic. Even though research on algorithms using pretrained feature extractors with frozen weights (REMIND, SLDA, DSDM…) claim to address the problem of continual learning without further hypothesis, the previously presented experiments suggests that there is a hidden assumption that the features that differentiate the classes in the newly learned tasks will be the same that those in the tasks used to train the pretrained model. To the best of our knowledge, no studies have addressed the continual-learning issue when the distribution of data drastically changes between tasks. In general, the standard procedure is to split a relatively homogeneous dataset (e.g. picture from the real world) into various tasks.

We suggest that truly agnostic continual learning should rely on operations that are applied directly to the input data, rather than on transformed representations from those data. This is because at a given time, as one cannot know which features of a class will be discriminative in the future.

## IV. Continual learning through dataset compression with OPRE

We have shown that the use of a pretrained feature extractor with frozen weights can be restrictive and implies an implicit hypothesis on the future data to come. Beyond the issue of an algorithm relying on a pretrained feature extractor, the question of how agnostic others algorithms are is worth asking. In particular, as previously discussed, all methods which are not extending their network at each task rely on a fixed-size model whose size is optimized for the complete dataset from which all the tasks are drawn. Moreover, all methods rely on hyperparameters that are empirically adjusted to the full training dataset or dataset of exemplars of the same nature. Further, these hyperparameters are not interpretable based on the data from which they are derived, as illustrated in II.

In this part, we explore continual learning through of a simple online compression algorithm with minimal and explicit hypothesis on the incoming data. The compression algorithm aims at reducing the redundancy between data; to assess if the compression algorithm has preserved the relevant information at test time a deep convolutional model is trained on the compressed data and used to make the prediction. The study focuses on comparison of final accuracy and total memory footprint.

We do not claim that OPRE completely resolves the problem of catastrophic interference during continual learning. Instead, we argue that its behavior provides insight into how information loss and agnosticity can be handled more systematically in continual learning. OPRE should be understood from an information-theoretic viewpoint rather than as a full CL solution: by enforcing a transparent and controlled simplification of the input space, it isolates the role of information preservation in settings where future tasks may be arbitrarily defined. This perspective complements existing CL methods by highlighting what aspects of performance can be attributed purely to the structure of the retained information, independently of model-specific priors.

## 1. Presentation of the OPRE compression algorithm and its use for CL

The idea behind OPRE is to eliminate the storage of "redundant" information, i.e., information that is common across images, while keeping information that are different across images. It can be viewed as a lossy-dictionary compression algorithm in which the elements of the dictionary are small square patches making up the images. The loss of information occurs when a patch from a new image is not stored because it differs only very minimally (via a minimum Euclidean distance) from one or more patches that are already stored. For example, for the image of a bird, a small blue patch from the sky would not be stored if another very similar blue patch from another image has already been stored.

Each incoming three-channeled image is divided into small square patches with no overlap between them, i.e. here a 3x32x32 image is split into 64 patches of size 3x4x4. When a new patch arrives, it is first discretized: its pixel values are quantized to L levels per channel before any comparison takes place. Subsequently, each discretized patch is compared with all of the already seen patches stored in "patch memory", which therefore only ever contains discretized patches. If the Euclidian distance from at least one stored patch is below a specified threshold $\epsilon$, then the new patch is considered to be redundant and is discarded. If no stored patch is found to be close to this patch, the patch is added to patch memory. Each patch is assigned an integer ID identifier that codes for its position in the patch memory. The representation of the whole image now consists of a set of IDs corresponding to the patches making up the image. This allows the system to produce a compressed reconstruction of the full image. These representations are then stored in an "compressed-image memory", which thus contain images represented as an array of ID referring to patches stored in "patch memory".

OPRE algorithm is summarized in Algorithm 1. Parallelization is not made explicit for clarity.

**Algorithm 1**: OPRE algorithm and associated prediction model training in a continual learning setting (without parallelization)

```
Input:  sequence of tasks (T_1,…,T_K), distance threshold ε,
      number of discretization levels L
Output: patch_memory, image_memory

1:  patch_memory ← ∅
2:  image_memory ← ∅
3:  for each task T_k do
4:     for each image in T_k do
5:        quantize the pixel values to L levels per channel
6:        split the quantized image into non-overlapping patches (e.g., 3×4×4)
7:        compressed_image ← ∅
8:        for each patch do
9:           find p ∈ patch_memory such that
```

```
10:             euclidean_distance(patch, p) < ε
11:          if such a patch exists then
12:             append ID(p) to compressed_image
13:          else
14:             add the discretized patch to patch_memory
15:             append ID(patch) to compressed_image
16:       add compressed_image to image_memory
17: train a randomly initialized prediction model for tasks (T_1,…,T_k) using images reconstructed from
patch_memory and image_memory.
```

Over the whole stream, this brute-force search performs $O(N \cdot M)$ distance computations—N the total number of patches, M the number of stored ones—so it is quadratic ($O(N^2)$) whenever a roughly constant fraction of patches is retained.

Note that, even though the algorithm can work in a totally sequential online manner, i.e. one sample after another, it is a highly parallelizable process: patches can be added by batch. In this case all the patches of the batch are discretized first; then the redundancy between elements of the batch has to be eliminated, before eliminating the redundancy between patches within the batch and previously stored patches.

In the proposed experiment, the training items are provided class by class, in a class incremental scenario. At test time, a new convolution neural network (CNN) model is trained starting with initial random weights, as in GDumb (Prabhu et al., 2020). While this could lead to unnecessary computation, we choose to start with a randomly initialized network because the focus of this paper is on classification performance and memory consumption. We also compare the performance of OPRE with random selection of sample in a finite buffer, as in GDumb (Prabhu et al., 2020), using the exact memory budget as in OPRE.

## 2. Hyperparameters' choice

OPRE has two hyperparameters, both defined directly in the input space. The number of discretization levels is fixed by a storage-alignment constraint: with L levels per channel, a 2×2×3 sub-patch (12 values) takes one of $L^{12}$ states and so requires $12 \cdot \log_2 L$ bits, and $L = 6$ is the largest integer for which this fits into a single 32-bit machine word ($6^{12} \approx 2.18 \times 10^9 < 2^{32}$, whereas $7^{12} \approx 1.38 \times 10^{10} > 2^{32}$); a full 3×4×4 patch, made of four such sub-patches, is thus encoded in exactly four 32-bit words (128 bits), and the resulting uniform grid has step $\Delta = 1/(L-1) = 0.2$ of the pixel dynamic range. The threshold $\varepsilon = 0.4$, by contrast, is applied to the already-discretized patches: every incoming patch is quantized to the L-level grid upon arrival, and the threshold then bounds the Euclidean distance, over the whole 48-value patch, between the new discretized patch and the stored ones

We set ε deliberately so that the two explicit sources of information loss in OPRE—the L-level discretization and the ε-deduplication—operate at the same scale. The displacement the pixel-level discretization level imposes is straightforward to quantify: modeling the rounding error as uniform on $[-\Delta/2, \Delta/2]$ and independent across the $d = 48$ coordinates, each has variance $\Delta^2/12$. Summed over the 48 coordinates, the squared Euclidean displacement encoding inflicts on a patch has expectation $d \cdot \Delta^2/12 = 4\Delta^2$, i.e. a root-mean-square displacement of $\Delta \cdot \sqrt{d/12} = 2\Delta = 0.4$ of the dynamic range. The perturbation the patch-level deduplication introduces is equally explicit: a discarded patch is displaced by at most ε onto the stored patch that absorbs it, and since deduplication operates on the L-level grid (every patch being discretized upon arrival), the displacement it actually applies is at most the largest grid distance strictly below ε. Setting ε = 0.4 = 2Δ—i.e. equal to the typical discretization displacement—thus places the deduplication perturbation (at most 0.35) and the discretization displacement (0.4) within the same order of magnitude: neither loss channel is negligible relative to the other, and each is characterized by a single interpretable number—the resolution Δ = 1/(L−1) for the encoding, the merging radius ε for the deduplication. This also pins ε from the method rather than from the data: ε is set as a fixed multiple of the grid step defined by L (here exactly 2Δ, the RMS discretization displacement), not tuned on any dataset.

## 3. Results

The chosen size of a colored patch is 3x4x4. The model used to evaluate the final accuracy obtained from compressed data is the 10 million-parameter CNN described in Appendix A. We used a CNN classifier without weight quantization, i.e. weight values are coded as 32-bits floating point values.

The final accuracies are given in the Table 2. The results of our algorithm are compared to Online LoRA (Wei et al., 2024) and DSDM (Pourcel et al., 2022), two state-of-the-art online continual-learning algorithms. We also implemented GDumb with a buffer of exact same size to compare our approach with this baseline in the exact same setting: same memory budget and training from scratch at test time.

| | Predictor training | CIFAR-10 | CIFAR-100 |
|---|---|---|---|
| No compression | Offline | 93.92 (0.14) | 74.27 (0.38) |
| OPRE | From scratch | 89.69 (0.25) | 63.99 (0.24) |
| GDumb | From scratch | 86.85 (0.32) | 56.62 (0.22) |
| Online LoRA | Online | - | 49.40 |
| DSDM | Online | 76.00 | 63.3 |

Table 2. Mean accuracy comparisons, in percentage. Predictor training is reported to indicate if the loss of accuracy results solely on the loss of information caused by the method or both from the loss of information and the training process. The results are averaged over five runs. SD in parentheses.

To evaluate the memory saved by removing patch redundancy, we compared the number of patches stored versus the number of total patches present in the original images. As there are 64 patches per image and 50,000 images in both CIFAR-10 and CIFAR-100, the total number of patches is 3.2M. The number of effectively stored patches is given in the Table 3.

| | CIFAR-10 | CIFAR-100 |
|---|---|---|
| No compression | 3.2 M | 3.2 M |
| OPRE | 1.816 M | 1.848M |

Table 3. Number of stored patches with and without OPRE.

The total memory costs are given in Table 4. The memory consumption of OPRE and the CNN trained on the compressed data consists of the memory load associated with the data itself (i.e. the sum of the size of the patch memory and the size of the image memory) and memory load associated with weights of the model trained on the compressed representation of the images.

| | CIFAR-10 (MB) | CIFAR-100 (MB) |
|---|---|---|
| No compression<br>(Data size \| model size) | 194.4 (153.60 \| 40.77) | 194.5 (153.60 \| 40.95) |
| OPRE<br>(Data size \| model size) | 82.6 (41.85 \| 40.77) | 83.3 (42.37 \| 40.95) |

| GDumb<br>(Data size \| model size) | 82.6 (41.85 \| 40.77) | 83.3 (42.37 \| 40.95) |
|---|---|---|
| Online LoRA | - | > 194.4 |
| DSDM | > 44.7 | > 44.7 |

Table 4 Memory cost associated with online continual learning algorithms. The sizes are given in megabytes. Only the cost of the feature extractor or pretrained model is considered for Online LoRA and DSDM.

We also apply OPRE to the random linearly generated, separated classes. Since the quantization grid and the threshold $\varepsilon$ are defined on pixel values in [0, 1], the $N(0, 1)$ pixels are first mapped into [0, 1] through the affine transformation $x \mapsto (x + 3)/6$, clipped to [0, 1]; the labels being computed on the original values, this $\pm 3\sigma$ mapping preserves linear separability and clips only $\approx$0.3% of the pixel values. On these rescaled data, with $\varepsilon = 0.4$, no redundancies were found and all the patches were conserved, yielding the setting described in Appendix B up to the L-level quantization of the pixel values; the data can then still be classified when using a CNN as classification model, as detail in Table 5.

| | Accuracy (%) | Memory cost (MB) |
|---|---|---|
| No compression | 80.63 (0.27) | 163.6 (122.88 \| 40.75) |
| OPRE | 78.90 (0.27) | 92.0 (51.20 \| 40.75) |

Table 5: Accuracy and memory cost on synthetic data, both with and without OPRE.

## 4. Discussion

The accuracies reported in Table 2 were obtained in a class-incremental scenario. The most informative comparison is with GDumb, which we ran at an identical memory budget and with the same train-from-scratch protocol: OPRE outperforms it by 2.84 points on CIFAR-10 (89.69 vs 86.85) and by 7.37 points on CIFAR-100 (63.99 vs 56.62). Because everything but the choice of what to store is held fixed, this gap isolates the contribution of patch-level redundancy elimination: at equal memory, retaining a diverse set of patches preserves more discriminative information than retaining whole raw images, and the advantage widens on the more diverse dataset. This matters all the more given that GDumb itself was reported to outperform several established replay methods, including MIR, ER, and iCaRL (Prabhu et al., 2020).

This advantage is accompanied by a qualitative difference in the nature of the information loss. GDumb discards whole images, chosen without regard to their content, so what is lost can be neither characterized nor bounded. OPRE quantizes the pixels of every incoming patch to L levels and discards only those discretized patches that fall within $\varepsilon$ of an already-stored patch: both losses are explicit, bounded, and defined directly in the input space. The type of information removed is therefore interpretable and can be related to meaningful aspects of the image—the discretization step, for instance, plays a role analogous to a contrast/color discrimination threshold. At an equal budget, OPRE is thus both more accurate and more transparent about what it forgets, which we regard as its main advantage over GDumb.

Among the reported online methods, OPRE clearly exceeds DSDM on CIFAR-10 (89.69 vs 76.0) and is on par with it on CIFAR-100 (64.0 vs 63.3), while outperforming Online-LoRA (49.4)—and it does so without any pretraining, whereas DSDM relies on a frozen ImageNet feature extractor and Online-LoRA on a transformer pretrained on ImageNet-1K.

The method most closely related to OPRE, AQM (Caccia et al., 2020), is not included in Tables 2 and 4, because neither its accuracy nor its memory can be placed on the same axis as the methods we report. Its CIFAR-10 accuracy (47.0%) is measured within a different evaluation benchmark, that of Aljundi et al. (2019), whose classifier—a small network trained for only a few epochs—caps at an i.i.d.-offline accuracy of 79.2%, well below the 93.9% reached by the large CNN we retrain to convergence. AQM's figure therefore sits on a markedly lower ceiling than ours and cannot be read against it. Its memory, in turn, is a budget fixed in advance whereas OPRE takes no budget as input, its footprint emerging in megabytes from the selection criterion (ε, L) and growing with the data. These differences reflect a difference in philosophy rather than a shortcoming of either method: for AQM, memory is a fixed budget to optimize against; for OPRE, it is an outcome of the data. When the amount and nature of future information are unknown, we regard the latter—memory as a consequence of the data rather than a constraint imposed on it—as the more agnostic stance.

One might worry that training OPRE's classifier from scratch is an unfair advantage. As far as data is concerned, it is not: the classifier sees only the memory accumulated online, with no access to future or held-out data, and—because that memory fully determines the classifier—an online OPRE could obtain the same accuracy by retraining on demand, the only cost being computation.

In memory (Table 4), OPRE's total footprint, including the classifier, falls between that of DSDM and Online-LoRA. It is far more frugal than Online-LoRA, whose >194 MB footprint is calibrated for large datasets and cannot be reduced when the data are simpler—on CIFAR-100, OPRE uses under half of that budget. DSDM is lighter than OPRE (its ResNet-18 backbone, >44.7 MB, is comparable in size to our classifier), but the two model costs are of a different kind: OPRE's classifier is trained from scratch on the stored data, adds no class information beyond what that data already encodes, and could be regenerated from the stored memory on demand—we include it only out of conservatism. DSDM's smaller footprint is qualitatively unsurprising: its pretrained extractor transforms each image into a compact representation, but—as argued in Section III—this distillation is exactly the non-agnostic, task-determined prior we set out to avoid, with no guarantee that the information a future task needs survives. OPRE's larger footprint is the price of refusing that prior, compressing only by an explicit, input-space criterion.

The pretrained backbone of DSDM or Online-LoRA, by contrast, cannot be regenerated from the stored representations: it is an irreducible, data-independent store that embeds the very prior on future data we set out to avoid. Additionally, OPRE's compressed data alone (≈42 MB) is already smaller than the DSDM feature extractor and about 3.7 times smaller than the original dataset.

Table 3 shows that 56.8% and 57.8% of patches are retained on CIFAR-10 and CIFAR-100 respectively: the ε-threshold removes slight less than half of all patches as redundant, a substantial gain. Retention is slightly higher on CIFAR-100, consistent with its greater diversity, though the effect is small enough that the two datasets are of comparable complexity for OPRE. On the synthetic random data, by contrast, no patch is redundant and all are kept. OPRE's memory thus adapts to the complexity of the data, in contrast to fixed-size models whose budget is set in advance.

About OPRE's generality, the arbitrary linearly-separable data of Section III, which a pretrained ImageNet feature extractor cannot classify, are preserved by OPRE: because the pixels are random rather than following the 1/f statistics of natural images (Field, 1987), no redundancy is found and every patch is kept—so the memory saving only stems from the pixel-level discretization, but the two classes remain learnable. Unlike feature-extractor methods, OPRE makes no assumption that the data resemble natural images.

Finally, in agnostic CL the total amount of information to store is unknown in advance, which makes online storage necessary; OPRE has this property, processing images one at a time and storing only a reference for each redundant patch. By construction, OPRE's behavior is also independent of task size, whereas most CL methods (Online-LoRA, DyTox, Co2L) degrade as tasks become smaller.

These results are best read from an empirical information-theoretic standpoint rather than as a leaderboard claim: only the storage is online, the classifier being retrained offline as in GDumb, which is precisely why the matched-budget GDumb comparison—not raw accuracy against methods with different protocols—is the meaningful control. AQM (Caccia et al., 2020) reaches markedly higher compression ratios and scales to larger and more diverse data, at the cost of low final accuracy; OPRE's contribution is the explicit, training-free, bounded handling of information loss, not the compression ratio itself.

# V. Conclusion

CL methods are evaluated according to their performance on restricted benchmarks, thereby enabling the comparison between methods. This type of evaluation is likely to favor algorithms designed for those benchmarks and which could not properly work beyond them. The rigorous evaluation of all the sets of possible data and tasks is obviously impossible, but we nevertheless demonstrate using synthetic data that methods relying on a pretrained feature extractor fail at learning all kinds of data, even if they perform well on data "close" to the data on which pretraining has been carried out. We think that the testing using the synthetic data proposed in our work could be used in the future to evaluate how algorithms can adapt to data that is radically different from the data they were trained on. Other CL methods add hyperparameters or increase the size of the model used. We argue that there is a need to control the information given *a priori* to CL algorithms, as a first step towards agnostic continual learning. To fulfill this requirement, we developed OPRE, an online dataset compression method with an explicit and interpretable criterion on which information in the input space to discard. This is designed to avoid an uncontrolled loss of information, as OPRE keeps track of every item input to the system, while eliminating the redundancies between them. Even if this method is not fully agnostic, since it still makes some assumptions about the input space, it goes a long way in this direction: it guarantees that, whenever the discriminative structure of the data survives the two explicit loss criteria, that structure is preserved in the stored memory—so that learning it remains possible in principle, the quality of the prediction model being the only remaining bottleneck, which we leave for future work. OPRE reaches classification accuracy comparable to the strongest reported online continual-learning methods while being far more memory-frugal than large fixed-size models such as Online-LoRA—whose footprint is calibrated for large datasets and cannot be reduced when the data are simpler—and, at an identical memory budget and training protocol, it outperforms the GDumb baseline. Above all, every bit of information OPRE discards is discarded by one of two explicit criteria defined directly on the input—L-level pixel quantization and ε-distance patch merging—and nothing is lost in an opaque or uncontrolled way. Such a transparent characterization of the information loss is rare in the continual-learning literature, where it is usually an implicit by-product of a learned representation. While OPRE is a deliberately simple model that relies on retraining a classifier from scratch, we believe this explicit handling of information loss makes such approaches worth developing further.

It leaves two main directions for improvement. The first concerns scalability: the search for redundant patches could be accelerated using pivot-based metric indexing (Chávez et al., 2001; Zhu et al., 2022), which uses anchor points to avoid comparing each patch against all elements in the patch memory, while the patch size—fixed here—could be made adaptive through a multi-resolution approach or by dynamically identifying the scale of redundancy within the data, allowing OPRE to scale to larger images. The second direction concerns prediction. A current limitation of our approach is that the prediction-model architecture is fixed; because the stored data and the classifier are cleanly separated, the architecture could instead be selected to match the compressed data, for instance by means of Neural Architecture Search. One could also perform prediction directly from the compressed data, for example using methods like DSDM (but without a feature extractor) that operate directly in the input space.

**Statement on the use of AI tools**

During the preparation of this work, the authors used a large language model to assist with language editing, to improve the clarity and consistency of the manuscript, and to verify bibliographic references. The tool was used under the authors' direction; all scientific content, experimental design, results, and conclusions are the authors' own. The authors reviewed and edited all AI-assisted output and take full responsibility for the content of this publication. No AI system is listed as an author.

**Authors contribution**

Conceptualization: RB, MM, RMF

Methodology: RB, RMF

Investigation: RB

Visualization: RB

Funding acquisition: MM

Project administration: MM, RMF

Supervision: MM, RMF

Writing – original draft: RB

Writing – review & editing: RB, MM, RMF

Words count : 10482

**Appendix A**

The CNN model used for prediction relies on 2D convolution layers, batch normalization layers, ReLU activation function, maximum pooling layers and global average pooling layers; this model is close to the one named simpleNet (Liu et al., 2023).

Its pytorch implementation is given in the following:

```
class SimpleNet(nn.Module):
  def __init__(self, num_classes=10):
    super(SimpleNet, self).__init__()
    self.features = nn.Sequential(
      nn.Conv2d(3, 64, kernel_size=3, padding=1),
      nn.BatchNorm2d(64),
      nn.ReLU(inplace=True),
      nn.Conv2d(64, 64, kernel_size=3, padding=1),
      nn.BatchNorm2d(64),
      nn.ReLU(inplace=True),
      nn.Conv2d(64, 64, kernel_size=3, padding=1),
      nn.BatchNorm2d(64),
      nn.ReLU(inplace=True),
      nn.MaxPool2d(2, 2),

      nn.Conv2d(64, 128, kernel_size=3, padding=1),
      nn.BatchNorm2d(128),
      nn.ReLU(inplace=True),
      nn.Conv2d(128, 128, kernel_size=3, padding=1),
```

```
    nn.BatchNorm2d(128),
    nn.ReLU(inplace=True),
    nn.Conv2d(128, 128, kernel_size=3, padding=1),
    nn.BatchNorm2d(128),
    nn.ReLU(inplace=True),
    nn.MaxPool2d(2, 2),

    nn.Conv2d(128, 256, kernel_size=3, padding=1),
    nn.BatchNorm2d(256),
    nn.ReLU(inplace=True),
    nn.Conv2d(256, 256, kernel_size=3, padding=1),
    nn.BatchNorm2d(256),
    nn.ReLU(inplace=True),
    nn.Conv2d(256, 256, kernel_size=3, padding=1),
    nn.BatchNorm2d(256),
    nn.ReLU(inplace=True),
    nn.MaxPool2d(2, 2),

    nn.Conv2d(256, 512, kernel_size=3, padding=1),
    nn.BatchNorm2d(512),
    nn.ReLU(inplace=True),
    nn.Conv2d(512, 512, kernel_size=3, padding=1),
    nn.BatchNorm2d(512),
    nn.ReLU(inplace=True),
    nn.Conv2d(512, 512, kernel_size=3, padding=1),
    nn.BatchNorm2d(512),
    nn.ReLU(inplace=True),
    nn.Conv2d(512, 512, kernel_size=3, padding=1),
    nn.BatchNorm2d(512),
    nn.ReLU(inplace=True),
    nn.AdaptiveAvgPool2d((1, 1)),
)
```

```
        self.classifier = nn.Sequential(
            nn.Flatten(),
            nn.Linear(512, num_classes)
        )

    def forward(self, x):
        x = self.features(x)
        x = self.classifier(x)
        return x
```

**Appendix B**: details about prediction of random linearly separated data

The data to classify rely on the generation of a random (32,32) tensor, following a normal distribution with mean 0 and variance 1, from which the orthogonal to the hyperplane separating the two classes is defined. This pattern is first extended to (32,32,3) by duplicating it three times with respect to the third dimension, then it is flattened to a 32x32x3 vector denoted $v_{hyperplane}$. The (32,32,3)-shaped images are then generated with random noise for each pixel, following a normal distribution with mean 0 and variance 1. Their classes are determined by the sign of the scalar product between the orthogonal vector to the hyperplane $v_{hyperplane}$ and their 32x32x3 flatten representations.

The tested feature extractor is ResNet-18. As the input size of images of this model is (224, 224, 3), the generated data are scaled up to this dimension (as is done when extracting features from CIFAR10 images).

We also checked that the categories into which this data fall can be predicted by a convolutional neural network (CNN) with the architecture described in Appendix A.

All of the training data are used at each epoch in this experiment. There is no continual learning in this case, as the goal is to determine whether information is lost during feature extraction.

The number of generated images is 50000 of which 80% (i.e. 40000) are used for training and the remaining 20% are used at test time. The multilayer perceptron (MLP) after the pretrained feature-extractor (Resnet18 pretrained on ImageNet1K) has two layers, with 256 and 2 neurons, respectively. The MLP to which the data are directly given as input has two layers of 512 and 2 neurons. The learning rate for each network is 0.001 and the optimizer is Adam. The batch size is 64.

Figure 2 illustrates the evolution of classification accuracy throughout the training process, assessed on both the training and test datasets. The models compared include a MLP with and without a feature extractor, as well as the previously described CNN. Notably, in contrast to the other configurations, the MLP with a feature extractor fails to demonstrate effective learning, as evidenced by consistently low accuracy - approaching random chance - and stable performance across both training and test datasets.

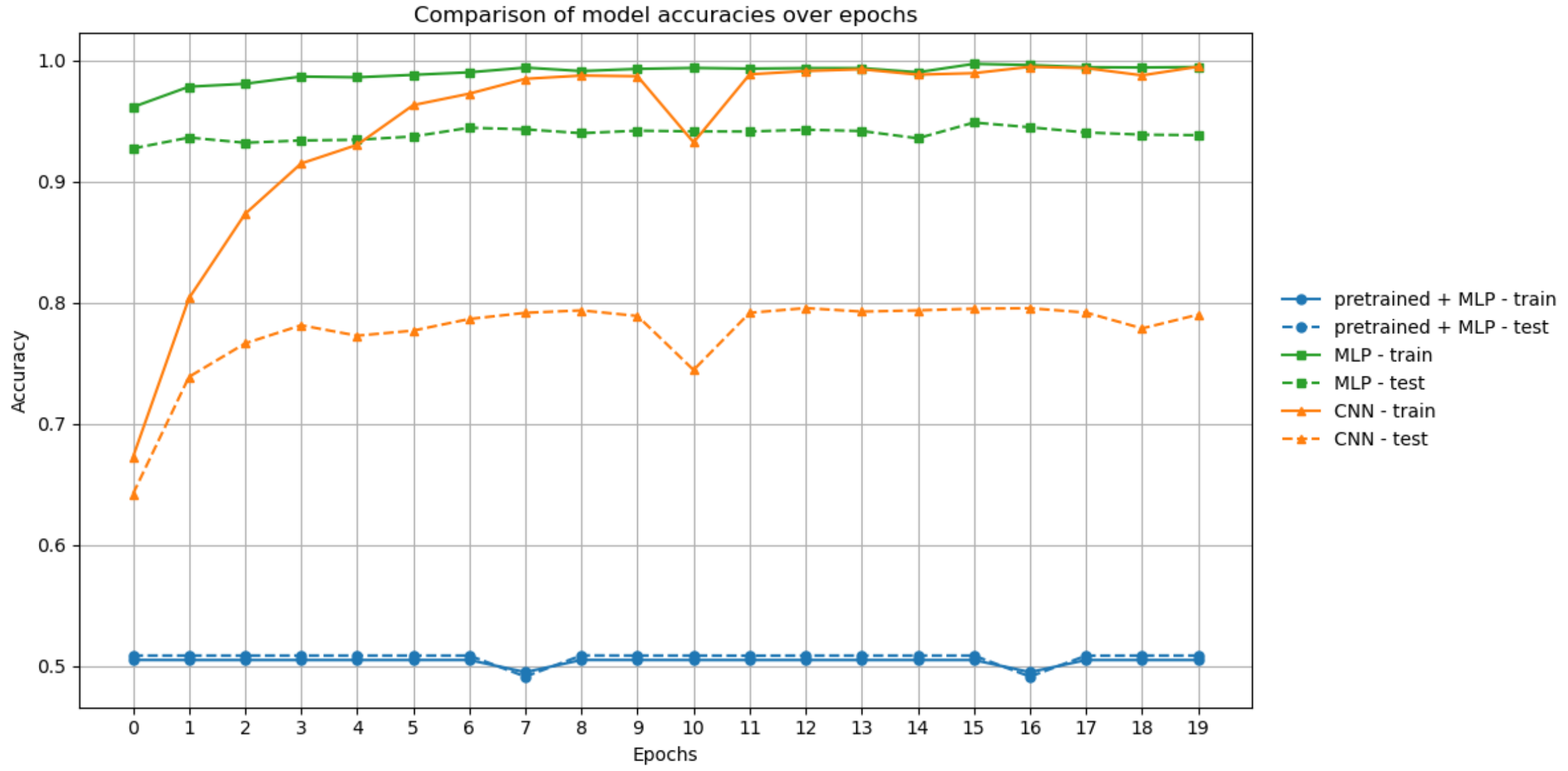

*Figure 2: Comparative analysis of model performance during training. The graph depicts the evolution of classification accuracy across epochs for three architectures: a multilayer perceptron (MLP) without feature extraction, a MLP preceded by a feature extractor, and a convolutional neural network (CNN) trained directly on raw input data. Accuracy is reported separately for the training and test datasets.*